\documentclass[journal,twoside,web]{ieeecolor}
\usepackage{generic}
\usepackage{cite}
\usepackage{amsmath,amssymb,amsfonts}
\usepackage{graphicx}
\usepackage[linesnumbered,ruled]{algorithm2e}
\usepackage{textcomp}
\usepackage{orcidlink}
\usepackage{multirow}
\usepackage{tabularx} 
\usepackage{booktabs}
\usepackage{hyperref,tikz}
\hypersetup{colorlinks=true,
            linkcolor=blue,
            anchorcolor=blue,
            citecolor=blue}

\def\BibTeX{{\rm B\kern-.05em{\sc i\kern-.025em b}\kern-.08em
    T\kern-.1667em\lower.7ex\hbox{E}\kern-.125emX}}
\begin{document}
\title{BRAU-Net++: U-Shaped Hybrid CNN-Transformer Network for Medical Image Segmentation}
\author{Libin Lan$^{\orcidlink{0000-0003-4754-813X}}$, \IEEEmembership{Member, IEEE}, Pengzhou Cai, Lu Jiang, Xiaojuan Liu$^{\orcidlink{0000-0002-9178-2006}}$, Yongmei Li, and Yudong Zhang$^{\orcidlink{0000-0002-4870-1493}}$, \IEEEmembership{Senior Member, IEEE}
\thanks{This paragraph of the first footnote will contain the date on which you submitted your paper for review. This work was supported in part by the Scientific Research Foundation of Chongqing University of Technology under Grants 0103210650 and 0121230235, and in part by the Youth Project of Science and Technology Research Program of Chongqing Education Commission of China under Grants KJQN202301145 and KJQN202301162. \textit{(Corresponding authors: Libin Lan; Yudong Zhang.)} }
\thanks{Libin Lan is with the College of Computer Science and Engineering, Chongqing University of Technology, Chongqing 400054, China,  with the College of Computer Science, Chongqing University, Chongqing 400044, China (e-mail: lanlbn@cqut.edu.cn).}
\thanks{Pengzhou Cai, and Lu Jiang are with the College of Computer Science and Engineering, Chongqing University of Technology, Chongqing 400054, China (e-mail: pengzhou.cai@stu.cqut.edu.cn; bdml\_jl@stu.cqut.edu.cn).}
\thanks{Xiaojuan Liu is with the College of Artificial Intelligence, Chongqing University of Technology, Chongqing 401135, China (e-mail: liuxiaojuan0127@cqut.edu.cn).}
\thanks{Yongmei Li is with the Department of Radiology, the First Affiliated Hospital of Chongqing Medical University, Chongqing 400016, China (e-mail: lymzhang70@aliyun.com). }
\thanks{Yudong Zhang is with the Department of Information Technology, Faculty of Computing and Information Technology, King Abdulaziz University, Jeddah 21589, Saudi Arabia, with the School of Computer Science and Engineering, Southeast University, Nanjing, Jiangsu 210096, China, and also with the School of Computing and Mathematical Sciences, University of Leicester, Leicester, LE1 7RH, UK (e-mail: yudongzhang@ieee.org).}}

\maketitle

\begin{abstract}
Accurate medical image segmentation plays an essential role in clinical quantification, disease diagnosis, treatment planning and many other applications. Both convolution-based and transformer-based u-shaped architectures have made significant success in various medical image segmentation tasks. The former can efficiently learn local information of images while requiring much more image-specific inductive biases inherent to convolution operation. The latter can effectively capture long-range dependency at different feature scales using self-attention, whereas it typically encounters the challenges of quadratic compute and memory requirements with sequence length increasing. To address this problem, through integrating the merits of these two paradigms in a well-designed u-shaped architecture, we propose a hybrid yet effective CNN-Transformer network, named BRAU-Net++, for an accurate medical image segmentation task. Specifically, BRAU-Net++ uses bi-level routing attention as the core building block to design our u-shaped encoder-decoder structure, in which both encoder and decoder are hierarchically constructed, so as to learn global semantic information while reducing computational complexity. Furthermore, this network restructures skip connection by incorporating channel-spatial attention which adopts convolution operations, aiming to minimize local spatial information loss and amplify global dimension-interaction of multi-scale features. Extensive experiments on three diverse imaging modalities datasets demonstrate that our proposed approach outperforms other state-of-the-art methods including its baseline: BRAU-Net under almost all evaluation metrics, which reveals the generality and robustness of our approach for multi-modal medical image segmentation tasks. The code and models are publicly available on \href{https://github.com/Caipengzhou/BRAU-Netplusplus}{GitHub}.
\end{abstract}


\section{Introduction}
\label{sec:introduction}

\IEEEPARstart{A}{ccurate} and robust medical image segmentation is an essential ingredient in computer-aided diagnosis systems, particularly in image-guided clinical surgery, disease diagnosis, treatment planning, and clinical quantification\cite{b1}, \cite{b2}, \cite{b3}. Medical image segmentation is usually considered to be essentially the same as natural image segmentation \cite{b4}, and that its corresponding techniques are often derived from that of the latter \cite{b5}. Common to the two communities is that they all take extracting the accurate Region of Interests (ROIs) of images as a study objective in a manual or automatic manner. Benefiting from deep learning techniques, the segmentation task in natural image vision has achieved an impressive performance. But different from natural image segmentation, medical image segmentation demands more accurate segmentation results for ROIs, e.g., organs, lesions, and abnormalities, to rapidly identify the ROI boundaries and exactly assess the level of ROI. This is because, in clinical practice, even a subtle segmentation error in medical images could degrade the user experience and increase the risk during subsequent computer-aided diagnosis \cite{b6}. Also, manually delineating the ROIs and their boundaries in various imaging modalities requires extensive effort that is extremely time-consuming and even impractical, and the resulting segmentation may be influenced by the preference and expertise of clinicians \cite{b7}, \cite{b45}. Thus, we believe that it is critical to develop intelligent and robust techniques to efficiently and accurately segment organs, lesion and abnormality regions in medical images.

Depending on the development of deep learning as well as the extensive and promising applications, many medical image segmentation methods which rely on convolution operations have been proposed for segmenting the specific target object in medical images. Among these approaches, the u-shaped encoder-decoder architectures like U-Net \cite{b8} and Fully Convolutional Network (FCN) \cite{b9} have become dominant in medical image segmentation. The follow-up various variants, e.g., U-Net++ \cite{b6}, U-Net 3+ \cite{b10}, Attention U-Net \cite{b11}, and 3D U-Net \cite{b12}, V-Net \cite{b13} have also been developed for 2D and 3D medical image segmentation of diverse imaging modalities, and made outstanding success in numerous medical applications such as multi-organ segmentation, skin lesion segmentation, and polyp segmentation. This indicates that Convolutional Neural Network (CNN) has a strong power to learn semantic information. But it often exhibits limitation in explicitly capturing long-range dependency due to the inherent locality of convolution operations. To tackle this limitation, some studies propose to enlarge receptive field by deep stacks of standard convolution operations or by dilated convolution operations \cite{b14}, \cite{b15}, \cite{b18}, or establish self-attention mechanisms relied on CNN features \cite{b16}, \cite{b17}. However, these methods can not remarkably improve the ability to model long-range dependency. 

Inspired by the recent success of applying transformer to Nature Language Processing (NLP) \cite{b19}, many studies attempt to incorporate transformer into vision domain \cite{b20}, \cite{b21}, \cite{b22}, \cite{b23}. These works have achieved consistent improvements on various vision tasks, which indicates that vision transformer has significant potential in the vision domain. However, vanilla transformer generally suffers from high computation cost and heavy memory footprint, which incurs a model efficiency problem in long-sequence scenarios. The improvement method most commonly used is introducing sparsity bias into the vanilla attention, i.e., adopting sparse attention instead of full attention to reduce computation complexity. The full attention needs to compute pairwise token similarity across all spatial locations, while the sparse attention allows each query token to just attend to a small number of key-value tokens instead of the entire sequence \cite{b24}, \cite{b25}. To this end, according to specific pre-defined patterns, some handcrafted static sparse attention methods are proposed, such as local attention \cite{b23}, dilated attention \cite{b26}, \cite{b27}, axial attention \cite{b28}, \cite{b31}, or deformable attention \cite{b55}. In medical image vision community, many studies also consider bringing transformer into medical image segmentation task, like nnFormer \cite{b29}, UTNet \cite{b30}, TransUNet \cite{b1}, TransCeption \cite{b3}, HiFormer \cite{b32}, Focal-UNet \cite{b33}, and MISSFormer \cite{b34}. However, to our knowledge, only several works consider introducing sparsity thought into this field, in which the representative works involve Swin-Unet \cite{b35} and Gated Axial UNet (MedT) \cite{b36}. But these sparse attention mechanisms merge or select sparse patterns in a handcrafted manner. Thus, these selected patterns are query-agnostic, which are shared by all queries. Applying dynamic and query-aware sparse attention mechanism to medical image segmentation still remain largely unexplored.

\begin{figure}
\centering
\includegraphics[width=1.0\linewidth, keepaspectratio]{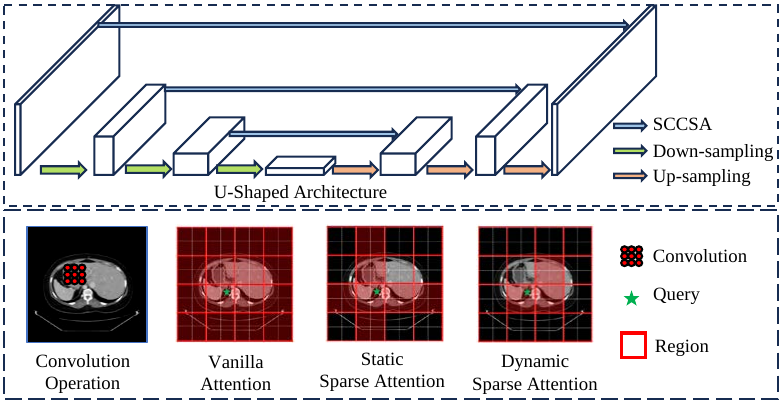}
\caption{Motivation. Due to the intrinsic locality of convolution operation as well as the high computation complexity of vanilla transformer, we consider incorporating sparse attention into U-shaped architecture, which can capture long-range dependency and reduce the computation cost to efficiently perform the medical image segmentation task. In practice, the main goal of using sparse attention mechanism is to ensure each query just attends to some most relevant key-value tokens. Since the tokens selected by static sparse attention are query-agnostic, we consider using query-aware, dynamic sparse attention mechanism in this work. Meanwhile, we consider restructuring skip connection with channel-spatial attention, which is implemented by convolution operation, aiming to amplify global dimension-interaction of multi-scale features.}
\label{fig0:motivation}
\end{figure}

\textbf{All these problems mentioned above motivate} us to explore a full-automatic advanced segmentation algorithm that can yield effective segmentation results relying on the nature of medical images, so as to benefit more image-guide medical applications (See Motivation Fig. \ref{fig0:motivation}). Inspired by the recent success of applying sparse attention \cite{b37} to vision transformer by BiFormer \cite{b24}, as well as using Swin Transformer \cite{b23} to build Swin-Unet \cite{b35} architecture, we propose, \textbf{BRAU-Net++}, to exploit the strong ability of transformer for multi-modal medical image segmentation. As far as we know, BRAU-Net++ is first hybrid model that considers incorporating dynamic sparse attention into a CNN-Transformer architecture. BRAU-Net++ is also developed from BRAU-Net \cite{b38}, which uses BiFormer block to build a u-shaped pure transformer network structure with skip connection for pubic symphysis-fetal head segmentation. Similar to Swin-Unet \cite{b35} and BRAU-Net \cite{b38}, the main components of the network structure include encoder, decoder, and redesigned skip connection. Both encoder and decoder are hierarchically built based on the core building idea of BiFormer \cite{b24}: bi-level routing attention, which can effectively model long-range dependency and save both computation and memory. Meanwhile, motivated by Global Attention Mechanism (GAM) \cite{b39}, we redesign the skip connection by incorporating channel-spatial attention, which is performed through convolution operations, aiming to minimize local spatial information loss and amplify global dimension-interaction of multi-scale features. Also, similar to \cite{b24}, \cite{b26}, \cite{b40}, \cite{b41}, the proposed architecture also utilizes depth-wise convolutions to implicitly encode positional information. Extensive experiments on three publicly available medical image datasets: Synapse multi-organ segmentation \cite{b56}, ISIC-2018 Challenge \cite{b42}, \cite{b43}, and CVC-ClinicDB \cite{b44} show that BRAU-Net++ can achieve a promising performance and robust generalization ability.

Our main contributions can be summarized as follows:
\begin{enumerate}
    \item We introduce a u-shaped hybrid CNN-Transformer network, which uses bi-level routing attention as core building idea to design the encoder-decoder structure, in which both encoder and decoder are hierarchically constructed, so as to effectively learn local-global semantic information while reducing computational complexity.
    \item We redesign the traditional skip connection using  channel-spatial attention mechanism and propose the \textbf{S}kip \textbf{C}onnection with \textbf{C}hannel-\textbf{S}patial \textbf{A}ttention (SCCSA), aiming to enhance the cross-dimension interactions on both channel and spatial aspects and compensate for the loss of spatial information caused by down-sampling.
    \item We conduct extensive comparative and ablative studies to thoroughly evaluate the effectiveness of our final BRAU-Net++ on three commonly used datasets: Synapse multi-organ segmentation, ISIC-2018 Challenge, and CVC-ClinicDB datasets. As a result, the proposed BRAUNet++ demonstrates a better performance than other state-of-the-art methods under almost all evaluation metrics.
\end{enumerate}

The remainder of this paper is organized as follows. Section \hyperref[sec:relatedwork]{\uppercase\expandafter{\romannumeral2}} reviews prior related works. Section \hyperref[sec:method]{\uppercase\expandafter{\romannumeral3}} specifies our method, main building blocks, and training procedure. Section \hyperref[sec: exp-settings]{\uppercase\expandafter{\romannumeral4}} introduces our experimental settings. Section \hyperref[sec:exp-results]{\uppercase\expandafter{\romannumeral5}} reports the experimental details and results. Section \hyperref[sec:discussion]{\uppercase\expandafter{\romannumeral6}}  gives some discussions and specifications regarding the experimental results and findings, and finally, Section \hyperref[sec:conclusion]{\uppercase\expandafter{\romannumeral7}} presents our conclusion.

\section{Related work}
\label{sec:relatedwork}
\subsection{U-Shaped Architecture}
\subsubsection{CNN-Based U-Shaped Architecture for Medical Image Segmentation}

Main techniques of this paradigm involve U-Net \cite{b8} and FCN \cite{b9}, as well as subsequent various variants \cite{b6}, \cite{b10}, \cite{b11}, \cite{b12}, \cite{b13}, some of which, e.g., U-Net++ \cite{b6}, UNet 3+ \cite{b10} and 3D-Unet \cite{b12}, V-Net \cite{b13} are introduced into 2D and 3D medical image segmentation communities, respectively. The distinct property of this paradigm is that u-shaped architecture is constructed based on convolution operations. While this paradigm has achieved remarkable success in many medical applications due to its excellent feature representation capability, this line of technique mainly employs a series of convolution and pooling operations to design its encoder and decoder, which limits its ability to capture long-range dependency. In our work, we do not use convolution operation to the encoder and decoder of u-shaped network, but consider just applying its power representation ability to skip connection so as to enhance the global dimension-interaction of multi-scale features. With respect to more works about U-Net and its variants applied for medical image segmentation, readers can refer to the related review literatures \cite{b47}, \cite{b48}.

\subsubsection{Transformer-Based U-Shaped Architecture for Medical Image Segmentation}

The vanilla transformer architecture was initially proposed for machine translation task \cite{b19}, and has become the de-facto standard architecture on various NLP tasks. The follow-up works have made more attempts to extend transformer to computer vision. More recently, researchers have also tried to develop pure transformer or hybrid transformer to perform medical image segmentation. In \cite{b35}, a pure transformer, i.e., Swin-Unet, is proposed for medical image segmentation, in which the tokenized patches from raw image rather than CNN feature map, are fed into the architecture to learn local global semantic information. In contrast to Swin-Unet \cite{b35}, TransUNet \cite{b1} is proposed as a hybrid CNN-Transformer model, which takes as input the tokenized patches from CNN feature map instead of raw image to train the network, so as to obtain both detailed spatial information and global context, which are helpful for achieving superior segmentation performance. Similar to TransUNet, both UNETR \cite{b49} and Swin UNETR \cite{b50} employ transformer encoder and convolutional decoder to generate segmentation maps. These works use either full attention or static sparse attention to compute pairwise token similarity. Different from these methods, we use dynamic sparse attention to select most related tokens, and take as input of network the tokenized patches from raw image, so that the information is not lost due to lower resolution. 

\subsection{Sparse Attention Mechanism}

Sparsity mechanism has been used to address the computational cost and memory footprint of vanilla attention mechanism, which can lead to efficient transformer. The early sparsity thoughts in NLP primarily involve simple modifications to self-attention, and these modifications are generally based on handcraft-designed predefined patterns \cite{b61}. For instance, sparse transformer \cite{b37} only attends to some tokens at fixed intervals by sparse connection patterns. Due to the promising potential of sparsity mechanism, sparse attention has also gained more attraction in vision transformer \cite{b23}, \cite{b25}, \cite{b26}, \cite{b27}, \cite{b28}. For example, Swin Transformer uses local attention in local window to achieve a linear computation complexity. But this local attention mechanism is also handcraft-designed. Subsequent studies have also presented various manually designed sparse attention mechanism, such as dilated attention \cite{b26}, \cite{b27} or cross-shaped attention \cite{b31}. More recently, efficient vision transformer based on dynamic sparsity has achieved great progress. In \cite{b51}, a dynamic token sparsity mechanism is used to prune a large number of uninformative tokens so as to achieve model acceleration, while the accuracy basically remains unchanged. In \cite{b25}, \cite{b24}, quad-tree attention and bi-level routing attention are proposed respectively. For the two methods, though the pattern of selecting the tokens to be attended to is different, they all achieve adaptive sparsity. In this work, we attempt to use bi-level routing attention as basic sparse block to build a u-shaped encoder-decoder architecture for medical image segmentation.

\subsection{Channel-Spatial Attention}

Attention mechanism has made great success in computer vision, in which both channel attention and spatial attention are two important directions. Channel attention mainly focuses on the information of channels. For instance, Squeeze-and-Excitation Network (SENet) \cite{b52} adaptively recalibrates channel-wise feature responses to enhance the discriminative ability of features and improve the generality performance of the network. On the other hand, spatial attention generally focuses on relevant spatial regions. For example, Spatial Transformer Network (STN) \cite{b53} can transform various deformation data to an appropriate, expected result to simplify inference, e.g., in that scenario requiring an attention mechanism, higher resolution input can be transformed to lower resolution one, so as to improve computational efficiency. Considering the combination of channel attention and spatial attention, Convolutional Block Attention Module (CBAM) \cite{b54} arranges the two attentions in a channel-first sequential manner to effectively fucus on important features. But this method suffers from information reduction and dimension separation, which results in losing global channel-spatial interactions. In this work, inspired by Global Attention Mechanism (GAM) \cite{b39}, we use channel-spatial attention to redesign skip connection, so as to enhance channel-spatial dimension-interactions and compensate for the spatial information loss due to down-sampling.

\section{Method}
\label{sec:method}

In this section, we will give a detailed specification of our proposed approach. We start by briefly summarizing the \textbf{B}i-Level \textbf{R}outing \textbf{A}ttention (BRA) thought. We then introduce the BiFormer block built on this BRA thought, which is the main building block of our overall architecture. Also, we describe the main compenonts in sequence, including encoder, bottleneck, decoder, and \textbf{S}kip \textbf{C}onnection \textbf{C}hannel-\textbf{S}patial \textbf{A}ttention (SCCSA) module. Finally, we specify the overall architecture of the proposed BRAU-Net++ and its loss function and training procedure.

\subsection{Preliminaries: Bi-Level Routing Attention}

The Bi-level Routing Attention (BRA) is a dynamic, query-aware sparse attention mechanism. Its core idea is to remove the most irrelevant key-value regions in a coarse-grained region level, and only remain a few most relevant ones used for a fine-grained token level. In this level, a token-to-token attention is performed. Compared with other handcrafted static sparse attention mechanism \cite{b23}, \cite{b31}, \cite{b55}, the BRA is easy to model long-range dependency. On this point, it is similar to vanilla attention. But the BRA has a much lower complexity of $O((HW)^{\frac{4}{3}})$ than vanilla attention, which has a complexity of $O({(HW)^2})$ \cite{b24}.

\subsubsection{Region Partition and Linear Projection}

By dividing a 2D input feature map $\textbf{X} \in {\mathbb{R}^{H \times W \times C}}$ into $S$$\times$$S$ non-overlapped regions, the feature dimension  ${\frac{HW}{S^2}}$ of each region can be obtained. Correspondingly, the query, key, value $ {\textbf{Q}, \textbf{K}, \textbf{V}} \in {\mathbb {R}^{{S^2} \times {\frac{HW}{S^2}} \times C}}$ can be derived as linear projections of the resulting feature map $\textbf{X}^r \in {\mathbb {R}^{{S^2} \times {\frac{HW}{S^2}} \times C}}$:
\begin{equation}
    \textbf{Q} = \textbf{X}^r{\textbf{W}^q}, \textbf{K} = \textbf{X}^r{\textbf{W}^k}, \textbf{V} = \textbf{X}^r{\textbf{W}^v},
\end{equation}
where ${\textbf{W}^q}, {\textbf{W}^k}, {\textbf{W}^v} \in{{\mathbb R}^{C \times C}}$ are corresponding projection weight matrices for the query, key, value, respectively. 
 
\subsubsection{Region-to-Region Routing}

The process starts by calculating the average of \textbf{Q} and \textbf{K} for each region respectively, yielding region-level queries and keys, ${\textbf{Q}^r},{\textbf{K}^r} \in {{\mathbb R}^{{S^2} \times C}}$. Next, the region-to-region adjacency matrix, ${\textbf{A}^r} \in {{\mathbb R}^{{S^2} \times {S^2}}}$, is derived via applying matrix multiplication between ${\textbf{Q}^r}$ and transposed ${\textbf{K}^r}$. Finally, the key step is only keeping the top-$k$ most relevant regions for each query via a routing index matrix, ${\textbf{I}^r} \in {{\mathbb N}^{{S^2} \times k}}$, which is implemented by a row-wise top-$k$ operator: \texttt{topkIndex()}. The region-to-region routing can be formulated as:
\begin{equation}
{\textbf{A}^r} = {\textbf{Q}^r}{({\textbf{K}^r})^T},
\end{equation}
\begin{equation}
{\textbf{I}^r} = \operatorname{topkIndex}({\textbf{A}^r}).
\end{equation}

\subsubsection{Token-to-Token Attention}

\begin{figure}
\centering
\includegraphics[width=1.0\linewidth, keepaspectratio]{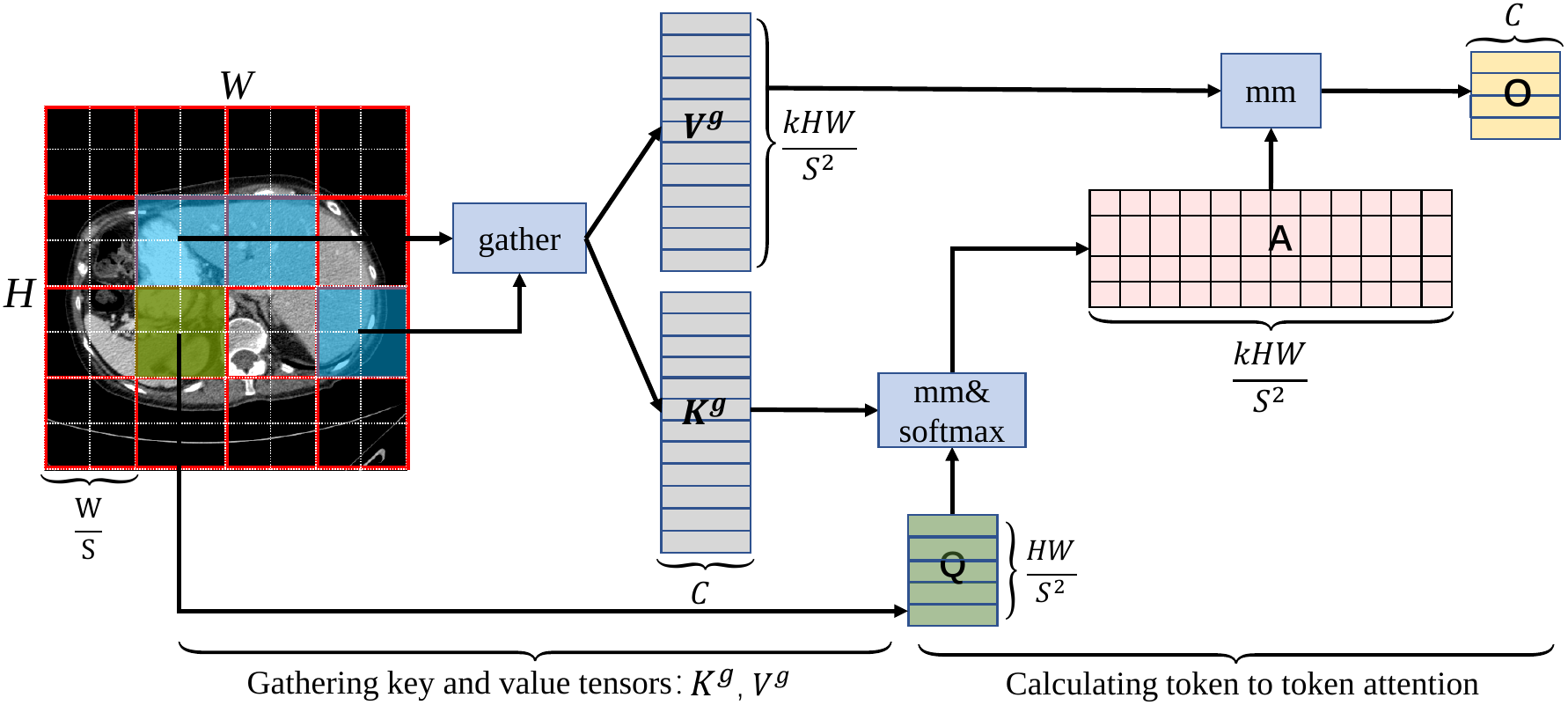}
\caption{Illustration of region gathering and token-to-token attention. By gathering the key and value tensors in routed regions, only GPU-friendly dense matrix multiplications are performed.}
\label{fig1:t2tattention}
\end{figure}

This process first needs to gather the routed regions due to the reason that these regions may be spatially scattered over the whole feature map. Then, a fine-grained token-to-token attention is applied to the gathered routed regions, in which a query in one region will attend to all key-value tokens in these gathered regions. This process is illustrated in Fig. \ref{fig1:t2tattention}, and can be formulated as follows:
\begin{equation}
{\textbf{K}^g} =\operatorname{gather}(\textbf{K},{\textbf{I}^r}), {\textbf{V}^g} =\operatorname{ gather}(\textbf{V},{\textbf{I}^r}),
\end{equation}
\begin{equation}
\textbf{O} = \operatorname{softmax}(\frac{{\textbf{Q}{{({\textbf{K}^g})}^T}}}{{\sqrt C }}){\textbf{V}^g} + \operatorname{LCE}(\textbf{V}),
\end{equation}
where ${\textbf{K}^g}, {\textbf{V}^g} \in {{\mathbb R}^{kHW \times C}}$ are gathered key and value tensors. The function LCE(·) is parameterized using a depth-wise convolution. Its kernel size is set to 5 in our all experiments.

\subsection{BiFormer Block}
\begin{figure}
\centering
\includegraphics[width=1.0\linewidth, keepaspectratio]{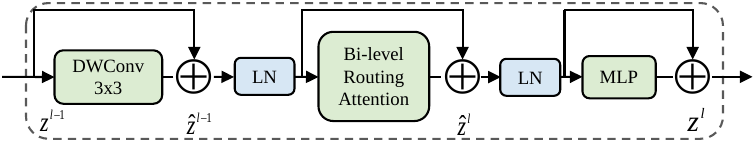}
\caption{Details of a BiFormer block.}
\label{fig3:bblock}
\end{figure}

The BiFormer block is built on this BRA mechanism. As illustrated in Fig. \ref{fig3:bblock}, It consists of three components: a 3$\times$3 depth-wise convolution, a BRA module, and a 2-layer Multi-Layer Perceptron (MLP) with expansion ratio $e$ = 3. Residual connection is used around each of the three components, followed by a layer normalization (LN). The BiFormer block can be formulated as:
\begin{equation}
{\hat{\mathbf{z}}^{l - 1}} = \operatorname{DW}({\mathbf{z}^{l - 1}}) + {\mathbf{z}^{l - 1}},
\end{equation}
\begin{equation}
{\hat{\mathbf{z}}^l} = \operatorname{BRA}(\operatorname{LN}({\hat{\mathbf{z}}^{l - 1}})) + {\hat{\mathbf{z}}^{l - 1}},
\end{equation}
\begin{equation}
{\mathbf{z}^l} = \operatorname{MLP}(\operatorname{LN}({\hat{\mathbf{z}}^l})) + {\hat{\mathbf{z}}^l},
\end{equation}
where ${\hat{\mathbf{z}}^{l - 1}}$, ${\hat{\mathbf{z}}^l}$ and ${\mathbf{z}^l}$ represent the outputs of the depth-wise convolution, BRA module and MLP module of the ${l^{th}}$ block, respectively.

\subsection{Encoder}
The encoder is hierarchically constructed by using a three-stage pyramid structure, in which the first stage consists of patch embedding layer and BiFormer blocks, and the second to third stages are composed of patch merging layer and BiFormer blocks. The number of BiFormer block is set to 2, 2, and 8 in each stage in sequence in our work. Our patch embedding layer employs two 3$\times$3 convolution layers to transform feature dimension ${\frac{HW}{16S^2}}$ of each region (for example, in stage 1, the resolution of feature map is $\frac{H}{4} \times \frac{W}{4}$, and the feature dimension of each region is 8$\times$8 = 64) to arbitrary dimension (i.e., channel, indicated as $C$). The patch merging layer uses a 3$\times$3 convolution layer to reduce spatial resolution of feature map by half while increasing dimension by 2$\times$. As illustrated in Fig. \ref{fig2:net}, in stage 1, the tokenized inputs with $S$$\times$$S$ regions (each region dimension is 64) and $C$ channels of each region are fed into the two consecutive BiFormer blocks to learn feature representation. In stage 2, the first patch merging layer performs a 2$\times$ down-sampling to make the resolution decreased to $\frac{H}{8} \times \frac{W}{8}$ and the feature dimension increased by 2$\times$ to 2$C$. In stage 3, this procedure is similar to that in stage 2, with resolution of $\frac{H}{16} \times \frac{W}{16}$ and 4$C$ dimension.

\subsection{Bottleneck}
Following Swin-Unet \cite{b35}, the bottleneck is composed of patch merging layer and BiFormer blocks, in which the number of BiFormer block is set to 2. The patch merging layer makes the dimension increased to 8$C$, i.e., the dimension of each region is 8$C$, and the resolution of feature map decreased to $\frac{H}{32} \times \frac{W}{32}$, i.e., each region size is 1$\times$1; that is, each region now is a pixel. The resolution and dimension of feature map passing through the two consecutive BiFormer blocks remain unchanged. 

\subsection{Decoder}
Similar to the encoder, the decoder is hierarchically built based on patch expanding layer and BiFormer block. Following Swin-Unet \cite{b35}, in the decoder, we adopt the patch expanding layer to up-sample the extracted deep features. The patch expanding layer decreases the feature dimension by half, and performs 2$\times$ up-sampling except for the last patch expanding layer, which performs 4$\times$ up-sampling to output the feature map of resolution $H \times W$, used to predict pixel-level segmentation. The number of BiFormer block is set as 8, 2, and 2, respectively, from stage 5 to stage 7. The feature map in each stage is divided into $S$$\times$$S$ regions, which are fed into follow-up BiFormer blocks.

\subsection{Skip Connection Channel-Spatial Attention (SCCSA)}
In contrast to only using a single attention mechanism, the combination of channel attention and spatial attention, especially the combination in a sequential manner, is helpful for improving the model's ability to capture important feature information \cite{b54}. Inspired by \cite{b39}, we consider applying a sequential channel-spatial attention mechanism to skip connection, and thus propose a skip connection channel-spatial attention, SCCSA for short. The SCCSA module can effectively compensate for the loss of spatial information caused by down-sampling and enhance global dimension-interaction of multi-scale features for each stage of the decoder, and thus enables the recovery of fine-grained details when generating output masks. As presented in Fig. \ref{fig2:net}{(b)}, the SCCSA module includes a channel attention submodule and a spatial attention submodule. Specifically, we first derive ${\mathbf{F}_1} \in {{\mathbb R}^{h \times w \times 2n}}$, via concatenating the output from both the encoder and the decoder. Then, the channel attention submodule uses a two-layer MLP with reduction ratio $e$ = 4, to magnify cross-dimension channel-spatial dependencies. In the spatial attention submodule, two 7$\times$7 convolution layers are used to focus on more spatial information, because it has relatively larger receptive field. For example, given the input feature map ${\mathbf{x}_1},{\mathbf{x}_2} \in {{\mathbb R}^{h \times w \times n}}$, the intermediate states ${\mathbf{F}_1},{\mathbf{F}_2},{\mathbf{F}_3}$, and the output ${\mathbf{x}_3}$ can be then formulated as:
\begin{equation}
{\mathbf{F}_1} = \operatorname{Concat}({\mathbf{x}_1},{\mathbf{x}_2}),
\end{equation}
\begin{equation}
{\mathbf{F}_2} = \sigma (\operatorname{FC}(\operatorname{ReLu}(\operatorname{FC}({\mathbf{F}_1}))) \otimes {\mathbf{F}_1},
\end{equation}
\begin{equation}
{\mathbf{F}_3} = \sigma (\operatorname{Conv}(\operatorname{ReLu}(\operatorname{BN}(\operatorname{Conv}({\mathbf{F}_2}))))) \otimes {\mathbf{F}_2},
\end{equation}
\begin{equation}
{\mathbf{x}_3} = \operatorname{FC}({\mathbf{F}_3}),
\end{equation}
where ${\mathbf{F}_2}$ and ${\mathbf{F}_3}$ are the output of channel and spatial attention submodule, respectively; $ \otimes $ and $\sigma $ denote element-wise multiplication and sigmoid activation function, respectively.

\subsection{Architecture Overview}
\begin{figure*}
\centering
\includegraphics[width=1.0\linewidth, keepaspectratio]{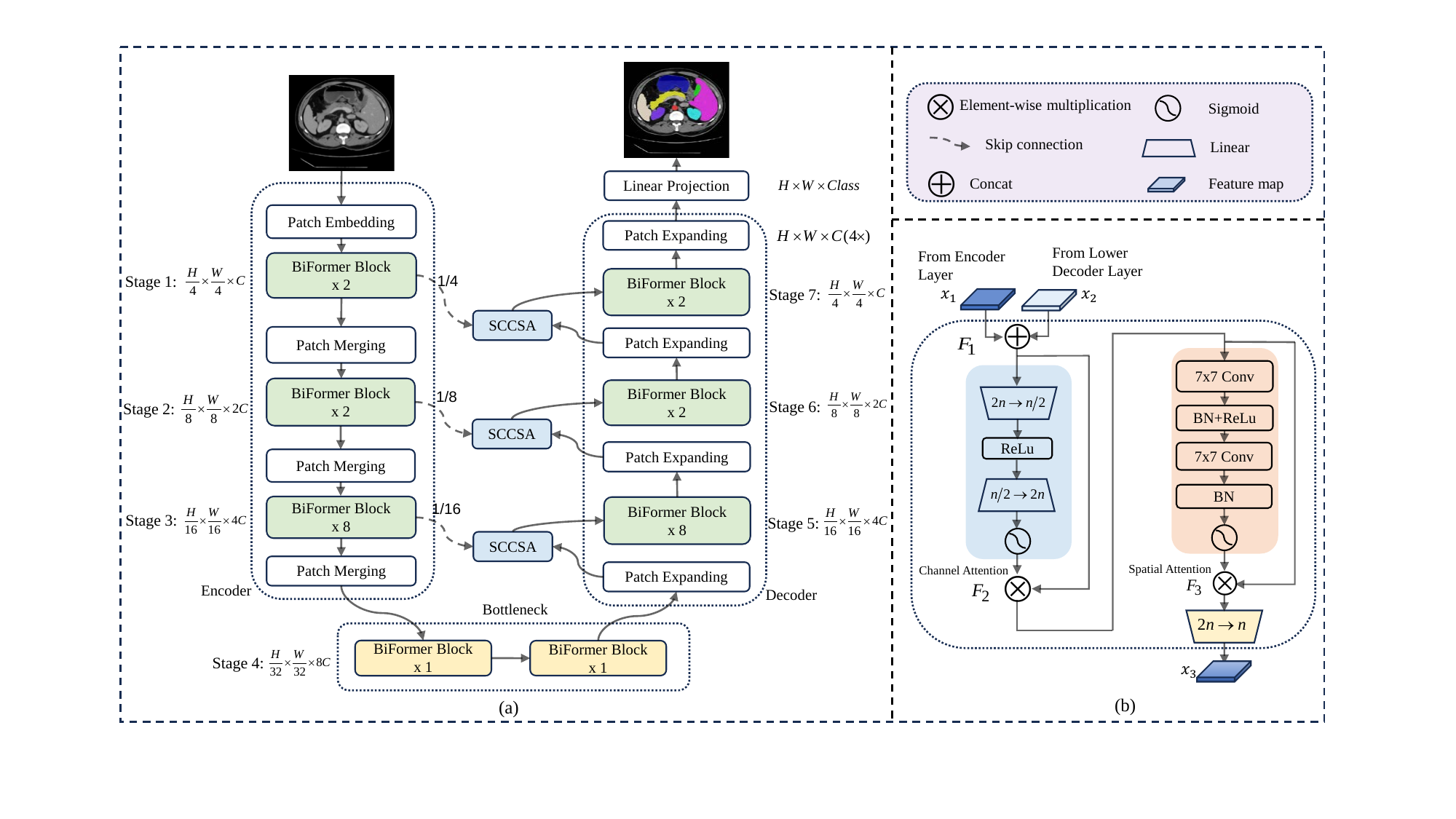}
\caption{(a): The architecture of our BRAU-Net++, which is a u-shaped hybrid CNN-Transformer network and uses a sparse attention mechanism: bi-level routing attention as core building idea to hierarchically design the encoder-decoder structure. (b): The skip connection channel-spatial attention (SCCSA) module, which is implemented mainly by convolution operation, aiming to enhance the ability of cross-dimension interactions from both channel and spatial aspects and compensate for the loss of spatial information caused by down-sampling.}
\label{fig2:net}
\end{figure*}

The BRAU-Net++ is composed of encoder, decoder, bottleneck, and SCCSA module, which forms a u-shaped hybrid network structure. The overall architecture of BRAU-Net++ is shown in Fig. \ref{fig2:net}{(a)}. On the top of network, a linear projection layer is applied on the feature maps of full resolution $H \times W $ to decrease their dimensions to number of class, which is used to predict the final pixel-level segmentation results. The core buildings of BRAU-Net++ are BiFormer block and SCCSA module. The network has 7 stages. Each stage from stage 1 to stage 7 has 2, 2, 8, 2, 8, 2, and 2 BiFormer blocks, respectively. The SCCSA module instead of traditional skip connection aggregates the features of different scales, which is implemented based on a global attention mechanism to minimize local spatial information loss and amplify global dimension-interaction of multi-scale features. The details of SCCSA module can be found in Fig. \ref{fig2:net}{(b)}. The whole network considers integrating the merits of self-attention and convolution to boost the ability to capture long-range dependency and to learn local information. Also, due to the dynamics and sparsity of bi-level rooting attention, the network has an advantage of low complexity.

\subsection{Loss Function}
In our experiments, we employ a hybrid loss function to train BRAU-Net++ and its various variants on Synapse dataset. This hybrid loss is the combination of dice loss ($\cal L$${_{dice}}$) and cross-entropy loss ($\cal L$${_{ce}}$), which is used to address the problems related to class imbalance. In addition, for the sake of optimization, we only use the dice loss to optimize BRAU-Net++ and its various variants on ISIC-2018 and CVC-ClinicDB datasets. The dice loss, the cross-entropy loss, and their hybrid loss ($\cal L$) are defined as follows:
 \begin{equation}
 {\mathcal{L}_{dice}} = 1 - \sum\limits_k^K {\frac{{2{\omega _k}\sum\nolimits_i^N {p(k,i)g(k,i)} }}{{\sum\nolimits_i^N {{p^2}(k,i) + \sum\nolimits_i^N {{g^2}(k,i)} } }}},
 \end{equation}
\begin{equation}
\begin{aligned}
\mathcal{L}_{ce} = & -\frac{1}{N}\sum_{i=1}^{N}(g(k,i) \cdot \log(p(k,i)) \\
& + (1 - g(k,i)) \cdot \log(1 - p(k,i))),
\end{aligned}
\end{equation}
\begin{equation}
\mathcal{L} = \lambda {\mathcal{L}_{dice}} + (1-\lambda) {\mathcal{L}_{ce}},
\end{equation}
where $N$ is the number of pixels, $g(k,i) \in (0,1)$ and $p(k,i) \in (0,1)$ indicate the ground truth label and the produced probability for class $k$, respectively. $K$ is the number of class,  and $\sum\nolimits_k {{\omega _k}}$ = 1 is weight sum of all classes. $\lambda$ is a weighted factor that balances the impact of $\mathcal{L}_{dice}$ and $\mathcal{L}_{ce}$. In our all experiments, The ${\omega _k}$ and $\lambda$ are empirically set as $\frac{1}{K}$ and 0.6, respectively. The training procedure of our BRAU-Net++ is summarized in Algorithm \ref{fig:alg}.
\begin{algorithm} 
\small
\caption{The training procedure of BRAU-Net++}
\label{fig:alg}
\SetKwInOut{Input}{input}\SetKwInOut{Output}{output}
\Input{Images $S$ = $\{{x_i}, i\in{{\mathbb N}}\}$, Masks $T$ = $\{{y_i^t}, i\in{{\mathbb N}}\}$} 
\Output{Model parameters}
	 \BlankLine 
    \For{$i = 0 \to batch$  $size $}{
    \emph{${x} = Patch$  $Embedding({x_i})$}\\
	 \For{$m = 0 \to num\_stage$}{  
	 	\For{$n = 0 \to num\_stage\_block$}{ 
                \emph{${x} = {x} + pos\_embed({x})$} \\   
                \emph{${x} = {x} + BRA({x})$} \\
                \emph{${x} = {x} + MLP({x})$}}
                \emph{${x_m} = Patch$  $Merging({x})$} \\
                \emph{${temp_m} = {x_m}$} 
 	 } 
        \For{$i = num\_stage-2 \to -1 $}{  
            \emph{${x_i} = Patch$  $Expanding({x})$} \\
            \emph{${x} = Concat({temp_i}, {x_{2-i}})$}\\
            \emph{${x} = SCCSA({x})$}\\

	 	\For{$j = 0 \to num\_stage\_block$}{
                \emph{${x} = {x} + pos\_embed({x})$} \\   
                \emph{${x} = {x} + BRA({x})$} 
                \\
                \emph{${x} = {x} + MLP({x})$} 
 	 	   }
       }
        \emph{${x} = Patch$  $Expanding$  4x$({x})$}\\
        \emph{${y_i^{out}} = Linear$  $Projection({x})$}\\
        \emph{Calculating the loss, $\cal L$ $\leftarrow$  $\lambda$ $\cdot$ $\cal L$${_{dice}}$(${y_i^{out}}$, ${y_i^t}$) $+ (1-\lambda) \cdot$ $\cal L$${_{ce}}$(${y_i^{out}}$, ${y_i^t}$)}\\
        \emph{Gradient back propagation, update parameters}
        }
    \end{algorithm}

\section{Experimental Settings}
\label{sec: exp-settings}
\subsection{Datasets}
We train and test the proposed BRAU-Net++ on three publicly available medical image segmentation datasets: Synapse multi-organ segmentation \cite{b56}, ISIC-2018 Challenge \cite{b42}, \cite{b43}, and CVC-ClinicDB \cite{b44}. The details about data split are presented in Table \ref{tab1:data}. All the datasets are related to clinical diagnosis, thus making their segmentation results crucial for the treatment of patients, and consist of the different modality images and their corresponding ground truth masks. These diverse imaging modalities datasets are deliberately selected to evaluate the generality and robustness of the proposed method. More details about these datasets are given as follows.
\begin{table}
\centering
\caption{Details of the medical segmentation datasets used in our experiments.}
\label{tab1:data}
\resizebox{1.0\linewidth}{!}
{
\begin{tabular}{l|ccccc}
\toprule
Dataset &Input Size & Total & Train & Valid & Test 
\\
\midrule
Synapse &224$\times$224 & 3379 & 2212 & 1167 & -
\\
ISIC-2018  &256$\times$256 & 2594 & 1868 & 467 & 259 
\\
CVC-ClinicDB &256$\times$256 & 612 & 490 & 61 & 61 
\\
\bottomrule
\end{tabular}
}
\end{table}

\subsubsection{Synapse Multi-Organ Segmentation Dataset}
The dataset used in our experiments includes 30 abdominal Computed Tomography (CT) scans from the MICCAI 2015 Multi-Atlas Abdomen Labeling Challenge, with 3,779 axial abdominal clinical CT images. Each CT volume involves 85--198 slices of 512$\times$512 pixels, with a voxel spatial resolution of ([0.54--0.54]$\times$[0.98--0.98]$\times$[2.5--5.0]) ${\operatorname{mm}^3}$. Following \cite{b1}, \cite{b35}, both training set and testing set consist of 18 (containing 2,212 axial slices) and 12 samples, respectively.

\subsubsection{ISIC-2018 Challenge Dataset}
The dataset in this work refers to the training set used for the lesion segmentation task in the ISIC-2018 Challenge, which contains 2,594 dermoscopic images with ground truth segmentation annotations. Five-fold cross validation is performed to evaluate the performance of model, and select best model to inference.

\subsubsection{CVC-ClinicDB Dataset}
The CVC-ClinicDB dataset is commonly used for polyp segmentation task. It is also the training dataset for the MICCAI 2015 Sub-Challenge on Automatic Polyp Detection Challenge. This dataset contains 612 images, which is randomly divided into 490 training images, 61 validation images, and 61 testing images. 

\subsection{Evaluation Metrics}
To evaluate the performance of the proposed BRAU-Net++, the average Dice-Similarity Coefficient (DSC) and average Hausdorff Distance (HD) are considered as evaluation metrics on 8 abdominal organs: aorta, gallbladder, spleen, left kidney, right kidney, liver, pancreas, spleen, and stomach, and only DSC is exclusively used on the evaluation of individual organ. Moreover, the mean Intersection over Union (mIoU), DSC, Accuracy, Precision, and Recall etc. are taken as evaluation metrics for the performance of models on both ISIC-2018 Challenge and CVC-ClinicDB datasets. Formally, the prediction can be separated into True Positive (TP), False Positive (FP), True Negative (TN), and False Negative (FN), and then DSC, IoU, Accuracy, Precision, and Recall are calculated as follows:
\begin{equation}
\operatorname{DSC} = \frac{{2 \times TP}}{{2 \times TP + FP + FN}},
\end{equation}
\begin{equation}
\operatorname{IoU} = \frac{{TP}}{{TP + FP + FN}},
\end{equation}
\begin{equation}
\operatorname{Accuracy} = \frac{{TP + TN}}{{TP + TN + FP + FN}},
\end{equation}
\begin{equation}
\operatorname{Precision} = \frac{{TP}}{{TP + FP}},
\end{equation}
\begin{equation}
\operatorname{Recall} = \frac{{TP}}{{TP + FN}}.
\end{equation}
HD can be described as:
\begin{equation}
\operatorname{HD}(Y,\hat Y) = \max \{ \mathop {\max }\limits_{y \in Y} \mathop {\min }\limits_{\hat y \in \hat Y} d(y,\hat y),\mathop {\max }\limits_{\hat y \in \hat Y} \mathop {\min }\limits_{y \in Y} d(y,\hat y)\},
\end{equation}
where $Y$ and $\hat Y$ are the ground truth mask and predicted segmentation map, respectively. $d(y, \hat y)$ denotes the Euclidean distance between points $y$ and $\hat y$.

\subsection{Implementation Details}
We train our BRAU-Net++ model and its various ablation variants on an NVIDIA 3090 graphics card with 24GB memory. We implement our approach using Python 3.10 and PyTorch 2.0 \cite{b57}. During training, we initialize and fine-tune the model on the above-mentioned three datasets, with the weights from BiFormer \cite{b24} pretrained on ImageNet-1K \cite{b58}, and considering space limits, also train the proposed model from scratch only on Synapse multi-organ segmentation dataset. On these resulting models, we conduct a serial of ablation studies to analyze the contribution of each component.

With respect to the Synapse multi-organ segmentation dataset, we resize all the images to the resolution of 224$\times$224, and train the model using stochastic gradient descent for 400 epochs, with a batch size of 24, learning rate of 0.05, momentum of 0.9, and weight decay of 1e-4. With regard to both ISIC-2018 Challenge and CVC-ClinicDB datasets, we resize all the images to resolution 256$\times$256, and train all the models using Adam \cite{b59} optimizer for 200 epochs, with a batch size of 16. We apply CosineAnnealingLR schedule with an initial learning rate of 5e-4. The data augmentations such as horizontal flip, vertical flip, rotation, and cutout with the probability of 0.25 are used to enhance data diversity. 

Other hyper-parameters are also empirically set. For example, the region partition factor $S$ is set as 7 and 8 according to the resolution of 224$\times$224 and 256$\times$256, respectively. The number of top-$k$ from stage 1 to stage 7 is set to 2, 4, 8, $S^2$, 8, 4, and 2, respectively, in which $S^2$ means using full attention.

\section{Experimental Results}
\label{sec:exp-results} 
In this section, we will elaborate on the comparisons of the proposed BRAU-Net++ with other state-of-the-art methods including CNN-based, transformer-based, and hybrid approaches of both on the Synapse multi-organ segmentation, ISIC-2018 Challenge, and CVC-ClinicDB datasets. Also, we conduct extensive ablation studies to analyze the effect of each component of our approach, in which the ablation study of SCCSA module is conducted on all three datasets, ablation studies of other components are only conducted on Synapse dataset.

\subsection{Comparison on Synapse Multi-Organ Segmentation}
The automatic multi-organ abdominal CT segmentation plays an essential role in improving the efficiency of clinical workflows including disease diagnosis, prognosis analysis, and treatment planning. So, we select this dataset to evaluate the performance of various methods. The comparisons of our proposal with previous state-of-the-art methods in terms of DSC and HD on Synapse multi-organ abdominal CT segmentation dataset are shown in Table \ref{tab2:synapse} with the best results in \textbf{bold}. The results of \cite{b32}, \cite{b60}, \cite{b33}, \cite{b34} are reproduced under our experimental settings according to the publicly released codes, while other results are directly from the respective published paper. Our BRAU-Net++ outperforms CNN-based methods and our baseline: BRAU-Net on both evaluation metrics by a large margin, which demonstrates that deeper hybrid CNN-Transformer model may be capable of modeling global relationships and local representations. Compared to both prevailing transformer-based methods: TransUNet \cite {b1} and Swin-Unet \cite{b35}, our BRAU-Net++ has a significant increase of 4.49\% and 3.34\% on DSC, and a remarkable decrease of 12.62mm and 2.48mm on HD, respectively. This indicates using bi-level routing attention as core building idea to design u-shaped encoder-decoder structure may be helpful for effectively learning global semantic information. More concretely, the BRAU-Net++ steadily beats other methods w.r.t. the segmentation of most organs, particularly for left kidney and liver segmentation. It can be seen from Table \ref{tab2:synapse} that the DSC value obtained by our method is highest, reaching up to 82.47\%, which shows that the segmentation map predicted by our method has a higher overlap with the ground-truth mask than other methods. One can also observe that we achieve a relatively low value (19.07mm) on HD compared to HiFormer and MISSFormer, which yields the best (14.7mm) and second-best (18.20mm) results, respectively. BRAU-Net++ just raises by 0.87mm on HD than MISSFormer, but has visibly increase of 4.37mm than HiFormer, which denotes that the ability of our methods to learn the edge information of target may be inferior to that of HiFormer. As a whole, Table \ref{tab2:synapse} shows that except for HiFormer and MISSFormer, the proposed BRAU-Net++ has significant improvements over prior works, e.g., performance gains range from 0.51\% to 12.2\% on DSC, and from 1.59mm to 20.63mm on HD, respectively. Thus, we believe that our approach has still a potential to obtain a relatively better segmentation result.

Also, one can see from Table \ref{tab2:synapse} that the number of learnable parameters of BRAU-Net++ is about 50.76M, in which SCCSA module yields about 19.36M parameters. But the performance of BRAU-Net++ with SCCSA module just slightly improves by 0.82\% on DSC than without SCCSA module. There is also a similar observation on HD. The effect of the number of parameters on performance will be discussed in the following section.

\begin{figure*}[htbp]
\centering
\includegraphics[width=1.0\linewidth, keepaspectratio]{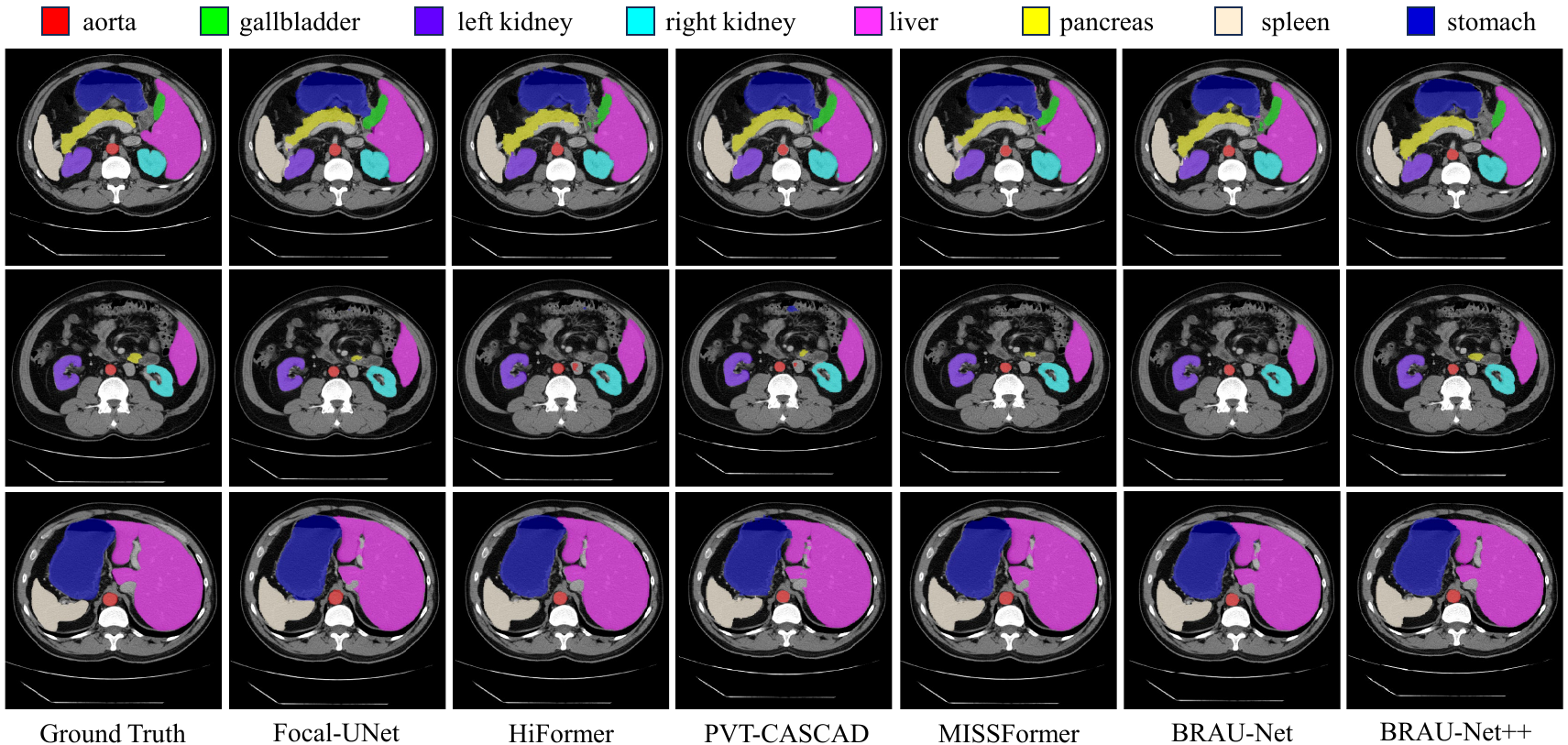}
\caption{Qualitative comparisons of our approach against other state-of-the-art methods on Synapse multi-organ segmentation dataset. Our BRAU-Net++ shows a relatively better visualization than other methods. Best viewed in color with zoom-in.}
\label{fig4:synapse}
\end{figure*}

Some qualitative results of different methods on the Synapse dataset are given in Fig. \ref{fig4:synapse}. It can be seen from Fig. \ref{fig4:synapse} that our method generates a smooth segmentation map for gallbladder, left kidney, and pancreas, which demonstrates that bi-level routing attention may excel at capturing the features of small targets, and the BRAU-Net++ can better learn both local and long-range semantic information, thus yielding a better segmentation result.

\begin{table*}
\centering
\caption{Quantitative results on Params, DSC, and HD of our approach against other state-of-the-art methods on Synapse multi-organ segmentation dataset. Only DSC is exclusively used for the evaluation of individual organ. The symbol $\uparrow$ indicates the larger the better. The symbol $\downarrow$ indicates the smaller the better. The best result is in \textbf{Blod}, and the second best is \underline{underlined}.}
\resizebox{1.0\linewidth}{!}{
\begin{tabular}{l|c|cc|cccccccc}
\toprule
Methods & Params (M)& DSC (\%) $\uparrow$  & HD (mm) $\downarrow$ & Aorta & Gallbladder & Kidney(L) &Kidney(R) &Liver & Pancreas & Spleen & Stomach\\
\midrule 
U-Net \cite{b8}&14.80   &76.85   &39.70 &\underline{89.07} &69.72 &77.77 &68.60 &93.43 &53.98 &86.67 &75.58
\\
Attention U-Net \cite{b11}&34.88  &77.77  &36.02 &\textbf{89.55} &68.88 &77.98 &71.11 &93.57 &58.04 &87.30 &75.75 
\\
BRAU-Net \cite{b38}& 33.30&70.27 &32.91 &78.51&61.69&72.94&67.90&93.14&40.88&84.42&62.68
\\
TransUNet \cite{b1}&105.28  &77.48   &31.69 &87.23 &63.13 &81.87 &77.02 &94.08 &55.86 &85.08 &75.62 
\\
Swin-Unet \cite{b35}&27.17  &79.13  &21.55 & 85.47 &66.53  & 83.28 &79.61  &94.29 & 56.58 &90.66  & 76.60
\\
HiFormer \cite{b32}&25.51  &80.39  &\textbf{14.70} & 86.21 &65.69  & 85.23 &79.77  &94.61 & 59.52 &90.99 & 81.08
\\
PVT-CASCADE \cite{b60}&35.28 &81.06 &20.23& 83.01 &\underline{70.59} & 82.23 &80.37  &94.08 & 64.43 &90.10  & \textbf{83.69}
\\
Focal-UNet \cite{b33}&32.40  &80.81  &20.66& 85.74 &\textbf{71.37}  &85.23 &\textbf{82.99}  &94.38 & 59.34 &88.49 & 78.94
\\
MISSFormer \cite{b34}&42.46  &\underline{81.96}  &\underline{18.20}& 86.99 &68.65  &85.21&82.00  &94.41 & \textbf{65.67} &\textbf{91.92} & 80.81
\\
\midrule
BRAU-Net++(w/o SCCSA)&31.40 &81.65  &19.46& 86.80 &69.73  &\underline{86.53} &\underline{82.24} &\underline{94.69}& 64.23 &89.69 & 79.26 
\\
BRAU-Net++ &50.76 &\textbf{82.47}  &19.07 & 87.95 &69.10  &\textbf{87.13} &81.53 &\textbf{94.71}  &\underline{65.17} &\underline{91.89}&\underline{82.26}
\\
\bottomrule
\end{tabular} }
\label{tab2:synapse}
\end{table*}

\subsection{Comparison on ISIC-2018 Challenge}
\begin{figure}
\centering
\includegraphics[width=1.0\linewidth, keepaspectratio]{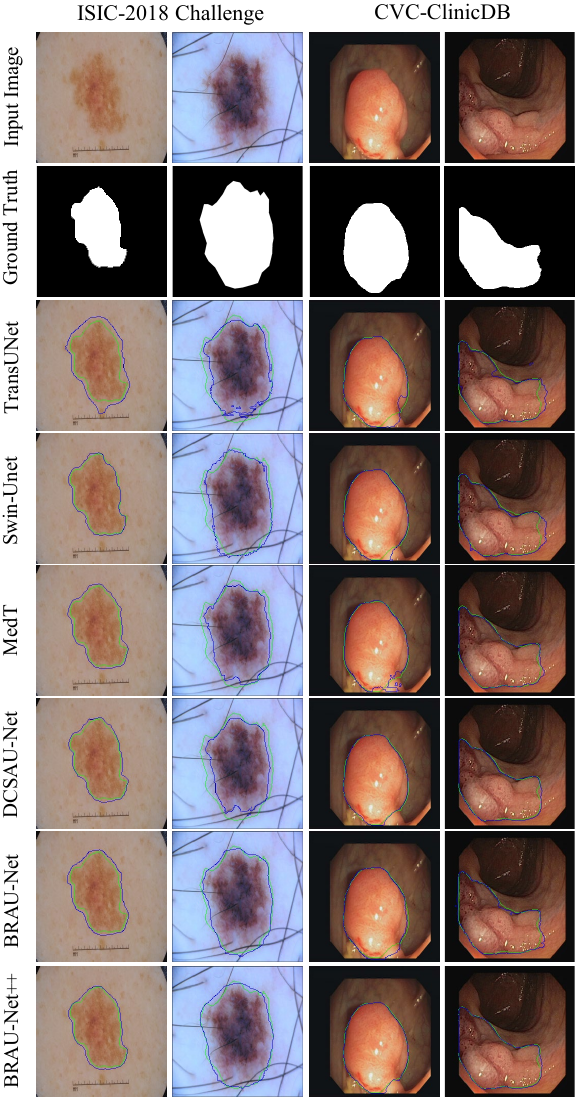}
\caption{Visualization comparisons of our approach against other state-of-the-art methods on both ISIC-2018 Challenge and CVC-ClinicDB datasets. Ground truth boundaries are shown in \textcolor{green}{green}, and predicted boundaries are shown in \textcolor{blue}{blue}.}
\label{fig5:ISICandCVC}
\end{figure}

It is well known that melanoma is a commonly occurring cancer, which if detected and treated in time, up to 99th-percentile of lives can be saved. So, an automated diagnostic tool for skin lesions is extremely helpful for accurate melanoma detection. We perform five-fold cross validation on ISIC-2018 Challenge dataset to evaluate the performance of our method, so as to avoid overfitting. We reproduce the results of all methods based on the publicly released codes. The quantitative and qualitative results are presented in Table \ref{tab3:ISIC} and in Fig. \ref{fig5:ISICandCVC} (left). Our method achieves mIoU of 84.01, DSC of 90.10, Accuracy of 95.61, Precision of 91.18, and Recall of 92.24, in which our method achieves the best performance in terms of mIoU, DSC, and Accuracy, and second-best result in terms of Precision and Recall. One can observe that the proposed BRAU-Net++ obtains improvements of 1.84\% and 1.2\% on mIoU over recently published DCSAU-Net \cite{b46} and preprinted BRAU-Net \cite{b38}, respectively. Also, our method achieves a recall of 92.24, which is more favorable in clinic applications. From the above analysis and Fig. \ref{fig5:ISICandCVC} (left), it can be evidently seen that BRAU-Net++ achieves better boundary segmentation predictions against other methods on ISIC-2018 Challenge dataset. The contours of segmented masks by BRAU-Net++ are closer to ground truth. 

\begin{table}
\centering
\caption{Qualitative results of different methods on ISIC-2018 Challenge dataset.}
\resizebox{1.0\linewidth}{!}{
\begin{tabular}{l|ccccc}
\toprule
\raggedright Methods & mIoU $ \uparrow$ & DSC $ \uparrow$ &Accuracy $ \uparrow$& Precision $ \uparrow$ &Recall $\uparrow$ 
\\
\midrule
U-Net \cite{b8} &80.21  &87.45 &95.21 &88.32 &90.60 
\\
Attention U-Net \cite{b11} &80.80  &86.31  &\underline{95.44} &\textbf{91.52} &89.01
\\
MedT \cite{b36} &81.43  &86.92 &95.10 &90.56 &89.93
\\
TransUNet \cite{b1}&77.05 &84.97 &94.56 &84.77 &89.85 
\\
Swin-Unet \cite{b35} &81.87  &87.43 &\underline{95.44} &90.97 &91.28
\\
BRAU-Net\cite{b38} &\underline{82.81} &\underline{89.32}&95.10&90.27&\textbf{92.25}
\\
DCSAU-Net\cite{b46}&82.17  &88.74 &94.75&90.93&90.98
\\
BRAU-Net++ &\textbf{84.01}  &\textbf{90.10} &\textbf{95.61} &\underline{91.18} &\underline{92.24}\\
\bottomrule
\end{tabular}}
\label{tab3:ISIC}
\end{table}

\subsection{Comparison on CVC-ClinicDB}
Before polyp has a potential to change into colorectal cancer, early detection can improve survival rate. This is of great significance to clinical practice. Therefore, we also select this dataset to validate the performance of model in our experiment.  The quantitative results are presented in Table \ref{tab4:CVC}. Our proposed method achieves best results on mIoU (88.17), DSC (92.94), Precision (93.84), and Recall (93.06), surpassing the second-best by 1.99\%, 1.27\%, 2.12\%, and 1.03\%, respectively. The qualitative results are shown in Fig. \ref{fig5:ISICandCVC} (right). One can see that the polyp masks generated by our approach closely match the boundaries and shape of ground truth.

\begin{table}
\centering
\caption{Qualitative results of different methods on CVC-ClinicDB dataset.} 
\resizebox{1.0\linewidth}{!}{
\begin{tabular}{l|ccccc}
\toprule
Methods & mIoU $\uparrow$ & DSC $\uparrow$ &Accuracy $\uparrow$& Precision $\uparrow$ &Recall $\uparrow$ 
\\
\midrule
U-Net \cite{b8}  &80.91  &87.22&98.45 &88.24 &89.35
\\
Attention U-Net \cite{b11} &83.54  &89.57 &98.64 &90.47 &90.10
\\
MedT \cite{b36} &81.47  &86.97 &98.44 &89.35 &90.04
\\
TransUNet \cite{b1}&79.95  &86.70 &98.25 &87.63 &87.34
\\
Swin-Unet \cite{b35} &84.85 &88.21 &98.72 &90.52 &91.13
\\
BRAU-Net \cite{b38} &77.45&83.64&97.96&84.56&84.20
\\
DCSAU-Net\cite{b46}&\underline{86.18}  & \underline{91.67} &\textbf{99.01} & \underline{91.72} & \underline{92.03}
\\
BRAU-Net++ &\textbf{88.17} &\textbf{92.94} &\underline{98.83}&\textbf{93.84}&\textbf{93.06}
\\
\bottomrule
\end{tabular}}
\label{tab4:CVC}
\end{table}

\subsection{Ablation Study}
In this section, we conduct an extensive ablation study to thoroughly evaluate the effectiveness of each component involved in BRAU-Net++. Specifically, we verify and analyze the impacts of SCCSA module on all three datasets, and the impacts of the number of skip connections and top-$k$, input size and partition factor $S$, as well as model scales and pre-trained weights only on Synapse dataset.

\subsubsection{Effectiveness of SCCSA Module}
The SCCSA module is an essential part of the proposed BRAU-Net++. It uses channel-spatial attention to enhance the cross-dimension interactions on both channel and spatial aspects, which is helpful for generating a more accurate segmentation mask. Table \ref{tab2:synapse} shows the results of BRAU-Net++ without and with SCCSA module (i.e, BRAU-Net++) on Synapse dataset. Compare with BRAU-Net++ without SCCSA, BRAU-Net++ achieves a better segmentation performance, increasing by 0.82\% on DSC and decreasing by 0.39mm on HD, respectively. But such a slight improvement comes at a cost: it brings a huge number of parameters into this model. One main reason may be that the combination of multi-scale CNN features with global semantic features learned by the hierarchical transformer structure can not significantly benefit the segmentation task. With respective to the exactly reasons, we intend to leave them as future work to further explore and analyze. The segmentation results on both ISIC-2018 Challenge and CVC-ClinicDB datasets are presented in Table \ref{tab5:performance}. One can see that adding SCCSA module into BRAU-Net++ model can achieve best results under almost all evaluation metrics. For example, SCCSA can help improve by 0.54\% on ISIC-2018 Challenge and by 0.8\% on CVC-ClinicDB w.r.t. mIoU metric, respectively. In addition, the number of parameters, floating point operations (FLOPs), and frames per second (FPS) are calculated to further investigate the effectiveness of this module. We can observe that SCCSA does not significantly harm FPS on the two datasets, particularly for ISIC-2018 Challenge dataset, which still seems to improve the FPS.

\begin{table*}
\centering
\caption{Ablation study on the impact of SCCSA module on both ISIC-2018 Challenge and CVC-ClinicDB datasets.} 
\resizebox{1.0\linewidth}{!}{
\begin{tabular}{l|c|ccc|ccccc}
\toprule
Dataset& Methods &Params (M) &FLOPs (G) &FPS & mIoU $\uparrow$ & DSC $\uparrow$ &Accuracy $\uparrow$& Precision $\uparrow$ &Recall $\uparrow$ 
\\
\midrule
\multirow{2}{*}{ISIC-2018 Challenge}&BRAU-Net++ (w/o SCCSA)&31.40 &11.12 &17.26&83.47  &89.75 &95.54 &91.01 &91.97 
\\
&BRAU-Net++ &50.76 &22.45 &29.84&\textbf{84.01} &\textbf{90.10} &\textbf{95.61}&\textbf{91.18}&\textbf{92.24} 
\\
\midrule
\multirow{2}{*}{CVC-ClinicDB}&BRAU-Net++ (w/o SCCSA)&31.40 &11.06 &15.95&87.37  &92.64 &\textbf{98.85} &\textbf{93.99} &92.01
\\
&BRAU-Net++ &50.76 &22.39 &15.56&\textbf{88.17} &\textbf{92.94} &98.83&93.84&\textbf{93.06}
\\
\bottomrule
\end{tabular}}
\label{tab5:performance}
\end{table*}

\subsubsection{Effectiveness of the Number of Skip Connections}
It has been witnessed that skip connection of u-shaped network can help improve finer segmentation details by using low-level spatial information \cite{b1}. This ablation mainly aims to explore the impact of the different numbers of skip-connections on the performance improvement of our BRAU-Net++. This experiment is conducted on Synapse dataset. The skip connections are added at the places of 1/4, 1/8, and 1/16 resolution scales, and the number of skip connections can be changed to be 0, 1, 2, and 3 through the combination of connections at different places, in which ``0'' indicates that no skip connection is added. Other added connections and their corresponding segmentation performance on DSC and HD are presented in Table \ref{tab6:skip}. We can observe that with the increase of the number of skip connections, the segmentation performance gradually increases, and best DSC and HD are achieved by adding the skip connections at all places of 1/4, 1/8, and 1/16 resolution scales. The finding is same as that in \cite{b1}, \cite{b35}. Thus, we adopt this configuration, of which the number of skip connections is 3, for our BRAU-Net++ to enhance the ability to learn precise low-level details. This may be main reason that BRAU-Net++ can capture the features of small targets.

\begin{table}
\centering
\caption{Ablation study of the number of skip connections on Synapse dataset.} 
\resizebox{1.0\linewidth}{!}{
\begin{tabular}{c|cccc|cc}
\toprule
\multirow{2}{*}[-0.8ex]{\# Skip Connection} &\multicolumn{4}{c|}{Connection Place}& \multirow{2}{*}[-0.8ex]{DSC $\uparrow$} & \multirow{2}{*}[-0.8ex]{HD $\downarrow$}
\\
\cmidrule{2-5}
&no skip& 1/4&1/8&1/16&&
\\
\midrule
0 & $\checkmark$&&&&76.40 & 28.36 
\\
1 & &$\checkmark$&&&78.56 & 26.14
\\
2 &&$\checkmark$&$\checkmark$&&81.16 & 22.67   
\\
3 &&$\checkmark$&$\checkmark$&$\checkmark$&\textbf{82.47} & \textbf{19.07}  
\\
\bottomrule
\end{tabular}}
\label{tab6:skip}
\end{table}

\subsubsection{Effectiveness of the Number of Top-k.}
Similar to \cite{b24}, as the size of the routed region gradually reduces at the following stage, we accordingly increase $k$ to maintain a reasonable number of tokens to attention. The results of ablation on the number of top-$k$ on Synapse dataset is showed in Table \ref{tab7:topk}, where the number of top-$k$ and tokens to attend in each stage of the network are listed. One can see that increasing the number of tokens near the bottom stages of encoder can seemingly improve the segmentation performance. That may be because the near bottom building blocks of network can capture low-level information e.g., edge or texture, which is essential for the segmentation task. Also, blindly increasing the number of tokens to attention may hurt the performance, which shows that explicit sparsity constraint can serve as a regularization to improve the generalization ability of model. This insight is similar to that in \cite{b24}.

\begin{table}
\centering
\caption{Ablation study of the number of top-$k$ on Synapse dataset.} 
\resizebox{1.0\linewidth}{!}{
\begin{tabular}{l|c|cc}
\toprule
\# top-$k$ & \# tokens to attend  & DSC $\uparrow$ & HD $\downarrow$ 
\\
\midrule
1,4,16,49,16,4,1 & 64,64,64,49,64,64,64 &81.83 & 23.92 
\\
2,8,32,49,32,8,2 & 128,128,128,49,128,128,128 & 81.74 & 23.21
\\
1,2,4,49,4,2,1 & 64,32,16,49,16,32,64& 82.03 & 21.54
\\
2,4,8,49,8,4,2 & 128,64,32,49,32,64,128  &\textbf{82.47} & \textbf{19.07}
\\
4,8,16,49,16,8,4 & 256,128,64,49,64,128,256  &82.08 & 20.09  
\\
\bottomrule
\end{tabular}}
\label{tab7:topk}
\end{table}

\subsubsection{Effectiveness of Input Resolution and Partition Factor S}
The main goal of conducting this ablation is to test the impact of input resolution on model performance. We perform three groups of experiments on 128$\times$128, 224$\times$224, and 256$\times$256 resolution scales on Synapse dataset, and report the results in Table \ref{tab8:resolution}. Following \cite{b24}, partition factor $S$ is selected as a divisor of the size of feature maps in every stage to avoid padding, and the images with different input resolutions should adopt different partition factors $S$. Thus, for the above three resolutions, we set corresponding partition factor as $S$ = 4, $S$ = 7, and $S$ = 8. It can be seen that keeping patch size same (e.g., 32) and gradually increasing the resolution scales (i.e., increasing the sequence length of tokens) can lead to the consistent improvement of model performance. It accords with the common sense that larger resolution images contain more semantic information, and thus boosting the performance. However, this is at the expense of much larger computational cost. Therefore, considering the computation cost, and to fair the comparison with other methods, all the experiments are performed by taking a default resolution of 224$\times$224 as the input.

\begin{table}
\centering
\caption{Ablation study of input resolution and partition factor $S$ on Synapse dataset. The symbol $\dag$ denotes the original resolution.}
\begin{tabularx}{1\linewidth}{ccccc|ccc|cccccc}
\toprule
 &&Image Size & & & &factor $S$ & &&  DSC $\uparrow$ && HD $\downarrow$ &&
\\
\midrule
&&128$\times$128 & & & &4 & &&77.99  & & 25.29 &&
\\
&&224$\times$224$^\dag$ &  && &7 & && 82.47  & & 19.07&&
\\
&&256$\times$256 & & & &8 & &&\textbf{82.61}  & &\textbf{18.56}&&
\\
\bottomrule
\end{tabularx}
\label{tab8:resolution}
\end{table}

\subsubsection{Effectiveness of Model Scale and Pre-trained Weights}
Similar to \cite{b1}, \cite{b35}, we discuss the effect of model scale on performance. Also, as we all known, the performance of transformer-based model is severely affected by pre-training. Thus, we consider providing four ablation studies on two different model scales, in which each model is trained from scratch and pre-trained respectively. The two different model scales are called ``tiny'' and ``base'' models, respectively. Their configurations and results on Synapse dataset are listed in Table \ref{tab9:scale}. One can see that the ``base'' model yields a better result. Particularly on HD evaluation metric, the result of the ``base'' model improves by 14.77mm compared to the ``tiny'' model. This suggests that the ``base'' model can achieve better edge prediction. Hence, we adopt the ``base'' model to perform medical image segmentation. Considering computation performance, we exploit the ``base'' model for all the experiments. 

\begin{table}
\centering
\caption{Ablation study of model scale and pre-trained weights on Synapse dataset.}
\resizebox{1.0\linewidth}{!}{
\begin{tabular}{ccc|cc|ccc}
\toprule
&Model Scale& & Channels &Params (M)  &DSC $\uparrow$&&HD $\downarrow$ 
\\
\midrule
&tiny w/o pre-t&&64 & 22.64&76.36&&34.04
\\
&tiny &&64& 22.64&79.39 &&33.84  
\\
&base w/o pre-t&&96 &50.76&78.48 && 23.84
\\
&base &&96&50.76 &\textbf{82.47} && \textbf{19.07}
\\
\bottomrule
\end{tabular}}
\label{tab9:scale}
\end{table}

\section{Discussion}
\label{sec:discussion}
\begin{figure*}
\centering
\includegraphics[width=1.0\linewidth, keepaspectratio]{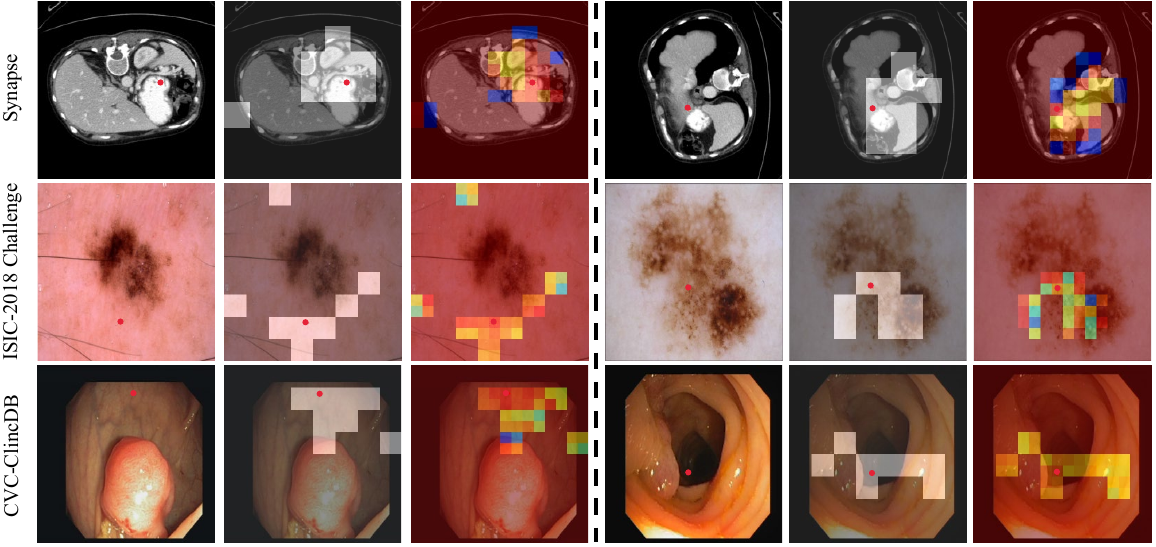}
\caption{Similar to \cite{b24}, visualization of attention maps on three datasets. For each dataset, we visualize a query position on the input image (left), corresponding routed regions (middle), and a final attention heatmap (right).}
\label{fig6:atten}
\end{figure*}

In this work, we show that the dynamic and query-aware sparse attention mechanism: bi-level rooting attention is effective on both reducing computational complexity and improving model performance. To further illustrate how the sparse attention works on medical image segmentation task, following \cite{b24}, we visualize routed regions and attention response w.r.t. query tokens. We adopt routing indices and attention scores, which are extracted from the final block of the $3^{rd}$ stage in the encoder, for this visualization. That is, these values are obtained from the feature map of $\frac{H}{{16}} \times \frac{W}{{16}}$ resolution, while the visualizations are presented in the images of original resolution. The qualitative results on Synapse multi-organ segmentation, ISIC-2018 Challenge, and CVC-ClinicDB datasets are shown in Fig. \ref{fig6:atten}. One can clearly see that the sparse attention mechanism can effectively find semantically most related regions, which indicates it is effective for the calculation and selection of sparse patterns of medical images. However, exploring other efficient sparse pattern computation methods is still necessary, and also the focus of our future work.

We perform a series of ablation studies to evaluate the contribution of each related component of BRAU-Net++, in which we propose SCCSA module to enhance the cross-dimension interactions of these features from stage $i$ in the encoder and from stage $7-i$ in the decoder on both channel and spatial aspects. The experimental results are encouraging under almost all evaluation metrics. However, one can see from Table \ref{tab2:synapse} that such a slight improvement comes at a cost of bringing a huge number of parameters. This is a shortcoming of our work. We believe main reason may be that the combination of multi-scale CNN features and global semantic features learned by the hierarchical transformer structure can not significantly benefit the segmentation task. In future work, we will focus on how to effectively address this problem. 

Three diverse imaging modalities datasets: Synapse multi-organ segmentation, ISIC-2018 Challenge, and CVC-ClinicDB, are deliberately chosen as benchmarks. The main reason of this choice is to evaluate the generality and robustness of the proposed method. Extensive experiments also reveal the generality of our approach for multi-modal medical image segmentation task.

\section{Conclusion}
\label{sec:conclusion}

In this paper, we propose a well-designed u-shaped hybrid CNN-Transformer architecture, BRAU-Net++, for medical image segmentation task, which exploits dynamic sparse attention instead of full attention or static handcrafted sparse attention, and can effectively learn local-global semantic information while reducing computational complexity. Furthermore, we propose a novel module: skip connection channel-spatial attention (SCCSA) to integrate multi-scale features, so as to compensate for the loss of spatial information and enhance the cross-dimension interactions. Experimental results show that our method can achieve state-of-the-art performance under almost all evaluation metrics on Synapse multi-organ segmentation, ISIC-2018 Challenge, and CVC-ClinicDB datasets, and particularly excels at capturing the features of small targets. For future work, we will focus on how to design more sophisticate and general architecture for multi-modal medical image segmentation task.

\section*{References}
\vspace{-1.5\baselineskip}

\end{document}